**3-D Volumetric Gamma-ray Imaging and Source Localization with a Mobile Robot – 18124**

Michael S. Lee[1], Matthew Hanczor[1], Jiyang Chu[2], Zhong He[2], Nathan Michael[1], Red Whittaker[1]
[1] The Robotics Institute, Carnegie Mellon University
[2] University of Michigan

**ABSTRACT**

Radiation detection has largely been a manual inspection process with point sensors such as Geiger-Muller counters and scintillation spectrometers to date. While their observations of source proximity prove useful, they lack the directional information necessary for efficient source localization and characterization in cluttered environments with multiple radiation sources. The recent commercialization of Compton gamma cameras provides directional information to the broader radiation detection community for the first time.

This paper presents the integration of a Compton gamma camera with a self-localizing ground robot for accurate 3D radiation mapping. Using the position and orientation of the robot, radiation images from the gamma camera are accumulated over a traversed path in a shared frame of reference to construct a consistent voxel grid-based radiation map. The peaks of the map at pre-specified energy windows are selected as the source location estimates, which are compared to the ground truth source locations. The proposed approach localizes multiple sources to within an average of 0.2 m in two 5 x 4 $m^2$ and 14 x 6 $m^2$ laboratory environments.

**INTRODUCTION**

There is substantial progress in the realm of perception for mobile robotic systems. Greater spatial understanding of the environment enable safe and more accurate navigation and localization, assisted by the recent commercialization of LIDAR sensors that provide robots with dense and accurate information about their surroundings [1, 2].

There is a similar breakthrough in radiation detection with the recent commercialization of Compton gamma cameras. Radiation detection, localization, and characterization is a common task in many nuclear activities, such as decommissioning and routine monitoring of facilities [3]. Geiger-Muller counters and scintillators, which are radiation detection sensors traditionally used for such tasks, can only convey point measurements of field strength [4]. These instruments require extensive observations over significant space and time to estimate source locations. Even with numerous readings these sensors can still fail to locate sources due to ambiguities that arise from non-unique combinations of source and environment configurations.

Gamma cameras determine the direction of incident photons via Compton imaging. The directional information of incident photons reduces the possible parameter space significantly and allows for practical use in general 3D environments. This work evaluates the ability of the Polaris-H gamma camera and a Compton imaging algorithm to localize source of specific isotopes in the environment through energy windowing.

To reliably locate radiation sources, Compton imaging depends on accurate pose information during detection. Robots are uniquely suited for accurate self-localization over long operations. By transforming gamma camera information over a robot trajectory into a shared frame of reference, a consistent radiation map can be built.

This paper presents the results of integrating a Compton gamma camera with a mobile ground robot to detect, localize, and characterize multiple gamma radiation sources in a lab environment. An overview of the proposed 3D source localization via volumetric gamma-ray imaging, quantitative evaluation to surveyed ground truth source locations, and qualitative evaluation in the context of robot-generated LIDAR maps are provided.

**RELATED WORK**

Point sensors such as Geiger-Muller counters and scintillators state spectrometers are ubiquitous in radiation detection due to their low cost. The two bodies of literature on using the observations of point sensors for radiation mapping and source localization are diverse, and are briefly surveyed below. But the ambiguity of point sensor observations forces each work to rely heavily on fundamentally limiting assumptions that, in turn, reduces its ability to generalize to arbitrary 3D environments.

One relevant body of work considers how to spatially interpolate or fit the point observations to a model to yield a radiation field map. Morelande et al. [5] obtain counts from a distributed sensor network and fit a Gaussian Mixture Model over the observed counts. They assume that the number of sources is known and leave model selection for future work. Minamoto et al. [6] build a radiation map of the ground by collecting radiation counts with a handheld dosimeter and finding a maximum a posteriori estimate over the distribution of point sources using the priors of inverse square law attenuation and Poisson count statistics. The strong priors practically limit this work to simple, open environments in which the inverse square law is valid. Martin et al. [7] fly a UAV equipped with a downward-facing rangefinder and a CZT spectrometer, constructing a 3D topological mesh and an overlay of the observed radiation data under an inverse square law assumption. While the inverse square law assumption is likely valid for aerial radiation mapping, this is likely insufficient for source localization as peaks of intensity measured by a point sensor may not be valid proxies for source locations.

Another relevant body of work for point sensors concerns source localization. Morelande et al. [8] estimates the source locations and intensities using a particle filter and the number of sources using the Bayesian Information Criterion, but it assumes an open, obstacle-free environment. Chin et al. [9] use a hybrid formulation of a particle filter and mean-shift techniques to localize multiple point sources by limiting the sensing range of the nodes of a sensor network and not associating particles to a particular source, allowing groups of nodes to separately localize sources in its neighborhood. Limitations of this method include the need for a densely populated sensor network, the possible difficulty in scaling the particle filter to 3D due to the curse of dimensionality, and the need to manually tune the neighborhood size. Towler et al. [10] use an Archimedian spiral search patterns to gather radiation count measurements from a RC helicopter, discover contour lines, and use a Hough transform to estimate the source positions and intensities for an arbitrary number of point sources. But this work assumes an open field and the Hough transform is based on overlapping circles that depends on the inverse square law. And lastly, Newaz et al. [11] also use a Gaussian mixture model to model the radiation field, and estimate the location and number of sources using variational Bayes inference of Gaussian mixtures. However, this method is also currently limited to 2D obstacle-free environments.

In contrast to point sensors, gamma cameras are able to directly provide the directional information of incident photons and enable new methods of radiation mapping. For example, Mihailescu et al. and Raffo-Caiado et al. [12, 13] overlay 2D gamma images from a Compton imaging- and coded aperture-based imagers, respectively, over a 3D model created by a laser scanner to create radiation-annotated point cloud for direction localization of hot spots without an assumption on the number or intensity of sources. A weakness in this method however, is that radiation can pass through objects and an overlay over the nearest object may incorrectly may label it as the source of the hotspot. Instead, movement of the gamma camera and triangulation of the observed directional rays enable a standalone 3D radiation map to be built, as is demonstrated in Haefner et al. [14]. This state-of-the-art 3D radiation mapping framework accumulates filtered back-projected images from an omnidirectional, hand-held CZT gamma camera in a voxel grid-based map of the environment built by RGB-D SLAM on a Kinect to localize multiple sources in a general 3D environment.

This paper extends the state of the art by deploying a mobile ground robot with an integrated Compton gamma camera for the remote characterization (i.e. 3D mapping and source localization) of nuclear environments. The hardware of the ground robot and gamma camera that enable localization and radiation imaging, respectively, are described in the following section. This is followed by an overview of the localization and imaging algorithms, experiments conducted, and the outcomes. The paper concludes with a discussion of results.

**METHOD - HARDWARE**

**Ground Robot**

All experiments were conducted using a custom-built skid-steer ground robot. A Velodyne LIDAR on the robot captures point clouds of the environment for map generation and continuous determination of robot pose. On-board visual cameras provide images for context and as a visual aid for the operator.

All tests were teleoperated by an operator with line of sight to monitor testing and to ensure accurate and repeatable trajectories. Two classes of motion were used in this testing: dwells at discrete, predetermined waypoints, and continuous driving. The dwell times for discrete tests was varied between experiments but for all continuous driving a speed of roughly 0.1 m/s was maintained.

The Polaris-H gamma camera was mounted on top of the robot to ensure an unobstructed view. The robot and mounted gamma camera used for these tests are shown in Figure 1.

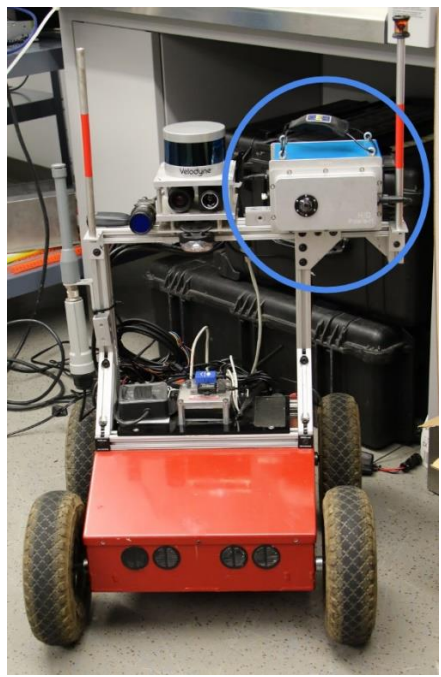

Figure 1 – Integration of a ground robot and gamma camera (circled).

**Radiation Detector**

The Polaris-H is an omnidirectional gamma camera that provides not only the count and energy of incident gamma photons but also their direction through Compton imaging. Compton imaging in turn relies on the physics of Compton scattering, in which a photon collides with a particle in its path, deposits a portion of its energy, and deflects in a new direction. Given the incident photon energy, deposited energies, and interaction locations, physics constrains the incident direction to a cone. As many of these Compton scattering events are detected, the accumulation of cones constrain and converge to the true direction of incidence, i.e. the source location.

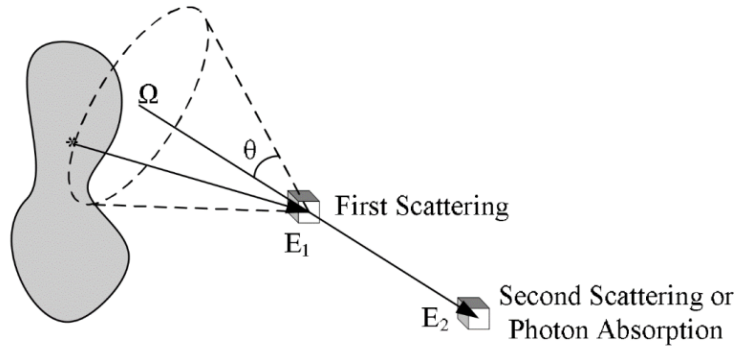

Figure 2 - Compton scattering physics that constrains direction of incident gamma ray to a cone given at least two interactions [15].

The heart of the Polaris-H is a 2 x 2 x 1.5 $cm^3$ CdZnTe crystal that consists of a pixelated array anode and a planar cathode that detect locations of interactions and the deposited energies, which is shown in Figure 3. The 2D interaction position is given by the triggered anode pixel, and the interaction height is given by the time that it takes the electron cloud generated by the photon to reach the anode. The aforementioned Compton scattering physics and the interaction locations and energy depositions are used to image the likely direction of the source.

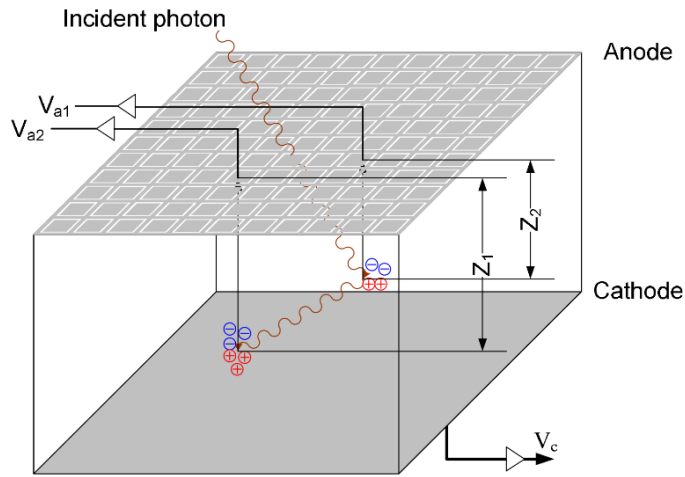

Figure 3 - CdZnTe detector for measuring interaction locations and energy deposition of incident photons [15].

**METHOD - ALGORITHMS**

**Localization and Mapping**

The Berkeley Localization and Mapping (BLAM) framework [16] provides localization and mapping using point cloud observations from the Velodyne LIDAR. BLAM consists of an inner loop and outer loop. In the inner loop, consecutive point clouds from the Velodyne are aligned using the Iterative Closest Point Algorithm (ICP) to provide a rough estimate of the instantaneous odometry. This odometry estimate is then used to seed the outer loop ICP between the most recent point cloud and the point cloud map maintained over time for robust localization. BLAM is built on a factor graph-based backend that records the history of point cloud observations and poses, and thus supports loop closures efficiently.

**Radiation Imaging**

Three algorithms are commonly used for radiation imaging. The first simply accumulates Compton cones in the imaging space and is thus called simple backprojection (SBP). Filtered backprojection (FBP) deconvolves the Compton cones with the spherical harmonic functions before accumulation to obtain a sharper image. The third is a statistical method called Maximum Likelihood Expectation Maximization (MLEM) which iteratively solves for the maximum likelihood gamma-ray source distribution in an expectation-maximization fashion given the observed radiation data. Though MLEM is slower than SBP or FBP due to its iterative nature, it achieves higher resolution and was thus selected as imaging algorithm for our experiments.

At a high level, MLEM works in two phases. In the first phase, an analytical inverse sensor model [15] is developed from the first principles of photon interaction physics (e.g. considering attenuation, Compton scattering, photoelectric absorption, etc). Given a set of observed interaction locations and energy depositions of an incident photon, the inverse sensor model assigns a probability to all possible directions of the photon's origin.

In the second phase, MLEM iteratively estimates the maximum likelihood source distribution by jointly considering the inverse sensor model and the observed data in an expectation-maximization fashion. The expectation (E) step associates the incident photon of each observed event to the likely direction of origin using the current estimate of the source distribution in the environment. The maximization (M) step then uses the photon-direction association from the expectation step to compute the next maximum-likelihood estimate of the source distribution. The optimal number of EM steps depends on the size of the imaging space, which depends on the number of discretization bins and the dimensionality of the imaging space (2D image or 3D map). This work ultimately considers a 3D imaging space to build a radiation map, representing it as a voxel grid (a voxel grid is the 3D equivalent of a 2D image pixel). The peaks of the voxel grids generated from radiation data within the pre-specified energy windows of the source isotopes of interest are selected as the source location estimates. Corresponding examples of 2D and 3D MLEM images from the same robot dwell point are shown below in Figure 4 and Figure 5.

In contrast to point sensors that struggle to reason about the location of the source due to ambiguities that arise from non-unique combinations of source and environment configurations, gamma cameras can determine the direction of the source regardless of the configuration of the source strength, source location, and the attenuation that occurs between the source and the detector, provided that the camera receives sufficient counts. Multiple radiation images taken at distinct locations can be used to triangulate the source location. This not only enables operation in a wider variety of contexts but also allows the peaks of the radiation maps to be taken as source locations with fewer assumptions on the source and environment configurations.

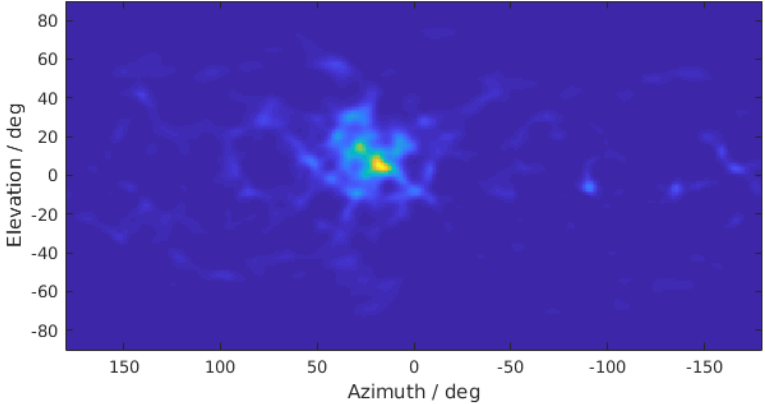

Figure 4 - 2D MLEM radiation image of a single source in front of the detector at 20 degrees azimuth and 5 degrees elevation.

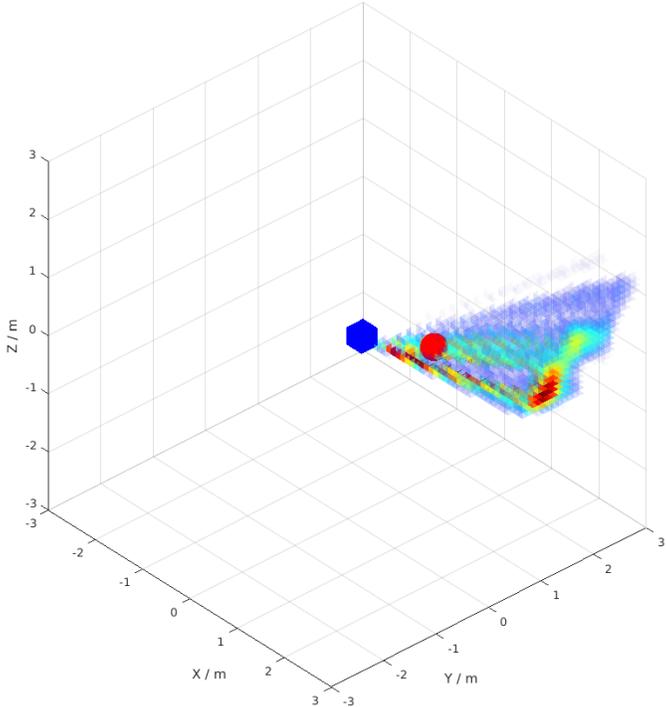

Figure 5 - Corresponding 3D MLEM radiation image to Figure 4, where the detector is shown as a blue cube, source is shown as a red sphere. Voxels are colored from red to blue, corresponding to the probability being along the source direction.

**EXPERIMENTS**

**Scenarios**

The experiments were conducted in two laboratory environments: a small uncluttered room with 5 x 4 $m^2$ meters of open driving space, and a much longer but cluttered space approximately 14 x 6 $m^2$ in size (shown in Figure 6).

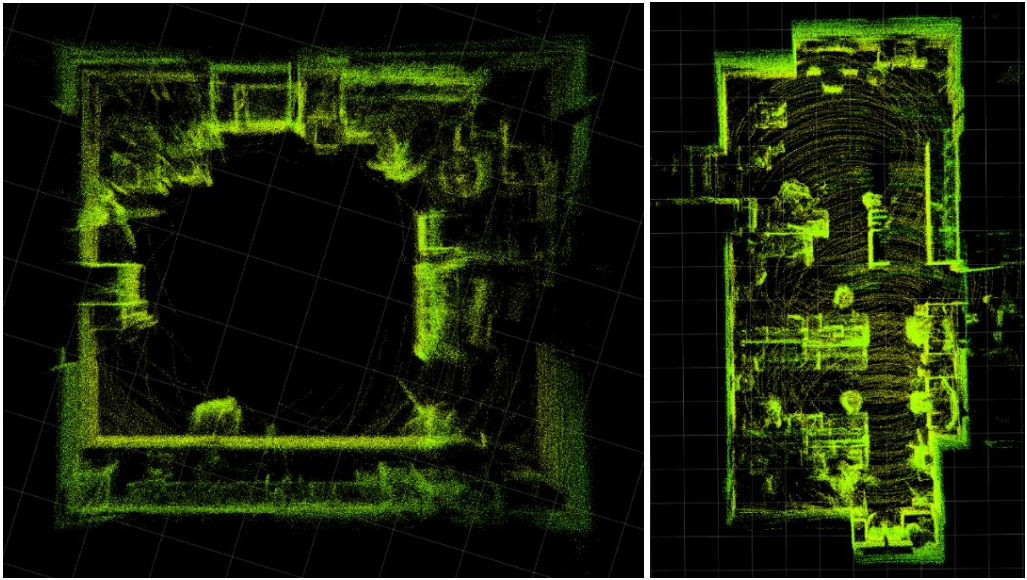

Figure 6 - Left: Small laboratory environment for testing (5 x 4 $m^2$).
Right: Large laboratory environment for testing (14 x 6 $m^2$).

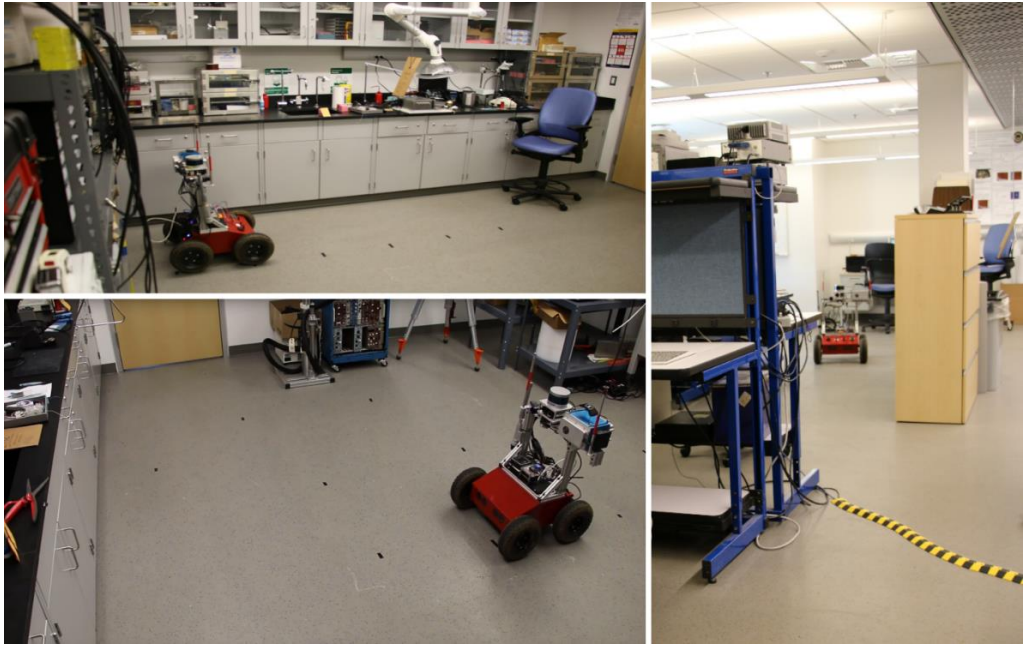

Figure 7 - Collage of experiments being carried out in the small (top left and bottom left) and large (right) laboratory environments.

A total of nine tests were performed in the two lab environments, where each test varied the robot's trajectory (i.e. straight line, spiral, or lawn-mower), motion (i.e. discrete dwells of varying lengths of time or continuous movement at a set speed), and surrounding point sources (i.e. a combination of Na-22, Cs-137, Co-60, and Ba-133 sources).

**Ground Truth**

Ground truth measurements of the source locations were recorded during testing to verify the accuracy of estimated source locations using a Leica Total Station surveyor, which has a stated accuracy of 1 mm.

**RESULTS**

Quantitatively, source localization results of sources using robot odometry and MLEM are compared to ground truth source locations for all tests. Select sources for two tests (test 5 and 7) were insufficiently imaged and thus the corresponding source localization errors are not included here. In test 5, the two Cs-137 sources in separate regions of the environment created a joint radiation map in which the peaks did not corresponding to the source locations. In test 7, Na-22, Co-60, and Ba-133 received insufficient imageable counts (i.e. events within the relevant energy window with 2 or more interactions that can be used for Compton imaging).

Details on the nine performed experiments can be found in Table I.

TABLE I. Experimental Details and Results

| Test | Location | Trajectory | Motion | Source Location(s) | Sources (uCi) | Imageable counts per source type | Source Localization Error (m) |
|---|---|---|---|---|---|---|---|
| 1 | Small room | Straight line | Discrete (6 dwells of 1 min each) | On a counter left of trajectory | Na-22 (61.28) | 1520 | 0.16 |
| 2 | Small room | Straight line | Discrete (6 dwells of 12 s each) | On a counter left of trajectory | Na-22 (61.28) | 261 | 0.17 |
| 3 | Small room | Straight line | Continuous | On a counter left of trajectory | Na-22 (61.28) | 63 | 0.36 |
| 4 | Small room | Straight line | Discrete (6 dwells of 1 min each) | On a counter left of trajectory behind attenuating material | Na-22 (61.28) | 767 | 0.10 |
| 5 | Large room | General | Discrete (10 dwells of 1 min each) | Each source on separate tables evenly spaced through the environment | Cs-137 (27.24) Na-22 (61.28) Co-60 (48.60) Cs-137 (100) | 131 1101 399 131 | n/a 0.32 0.08 n/a |
| 6 | Small room | Spiral | Continuous | At center of trajectory | Cs-137 (100) | 67 | 0.10 |

| 7 | Small room | Spiral | Continuous | Cs-137 at the center of trajectory, others distributed on a counter | Cs-137 (100) Na-22 (61.28) Co-60 (48.60) Ba-133 (82.11) | 67 4 0 18 | 0.33 n/a n/a n/a |
| 8 | Small room | Lawn mower | Continuous | Cs-137 at the center of trajectory, others distributed on a counter | Cs-137 (100) Na-22 (61.28) Co-60 (48.60) Ba-133 (82.11) | 62 48 17 89 | 0.03 0.12 0.20 0.43 |
| 9 | Small room | Lawn mower | Discrete (11 dwells of 1 min each) | Cs-137 at the center of trajectory, others distributed on a counter | Cs-137 (100) Na-22 (61.28) Co-60 (48.60) Ba-133 (82.11) | 106 206 63 161 | 0.05 0.23 0.23 0.21 |

Qualitatively, the estimated and ground truth source locations are placed in the context of the LIDAR map generated by the robot for quick spatial understanding of the source distribution in the environment for tests 1, 9, and 5. The count-energy spectrum of the radiation data collect during test 9 is included in Figure 10 to highlight the distinctive peaks in the energy windows used for the localization of sources.

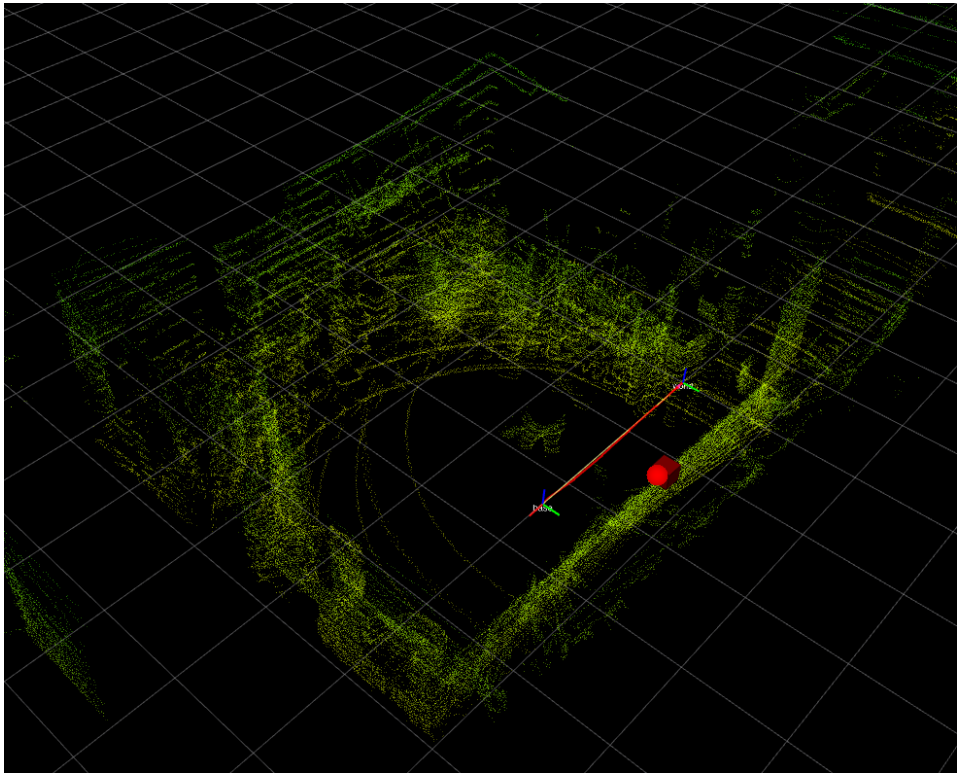

Figure 8 - Estimated location of Na-22 as a cube, with ground truth source location as a sphere, trajectory of robot in red, and contextual spatial map in green (test 1).

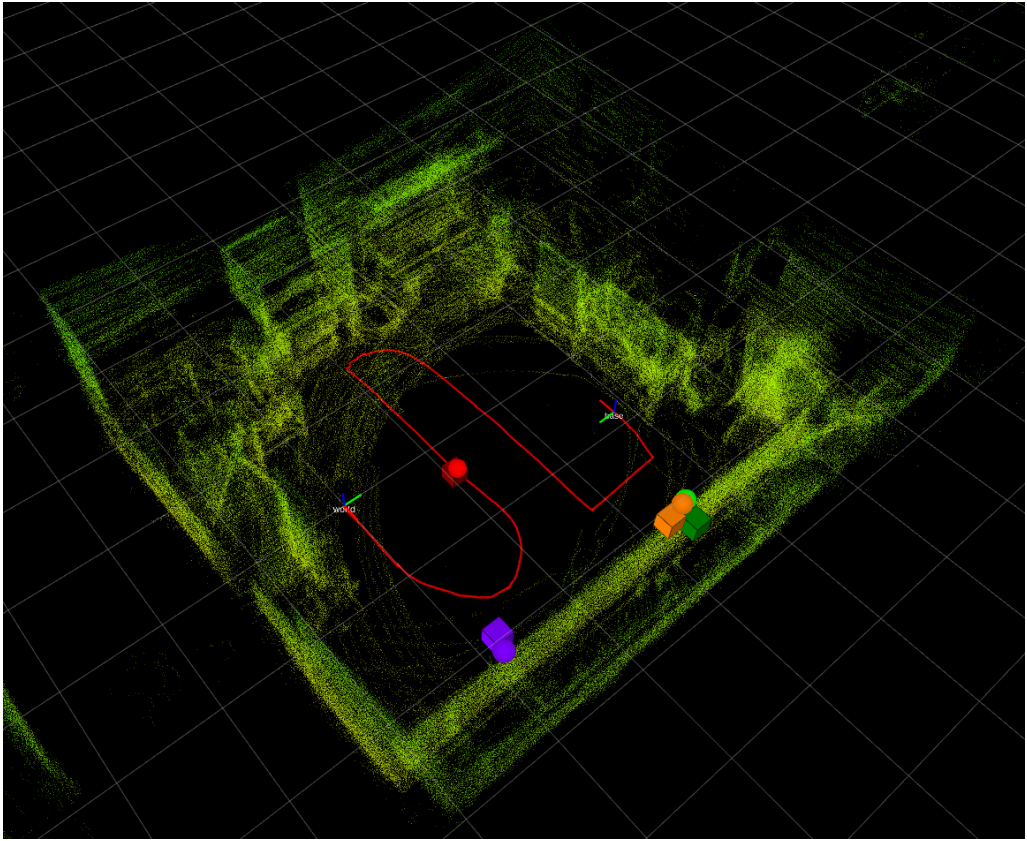

Figure 9 - Estimated locations of Ba-133 (purple), Cs-137 (red), Co-60 (orange), and Na-22 (green) as cubes, shown with ground truth source location as spheres, trajectory of robot in red, and contextual spatial map in green (test 9).

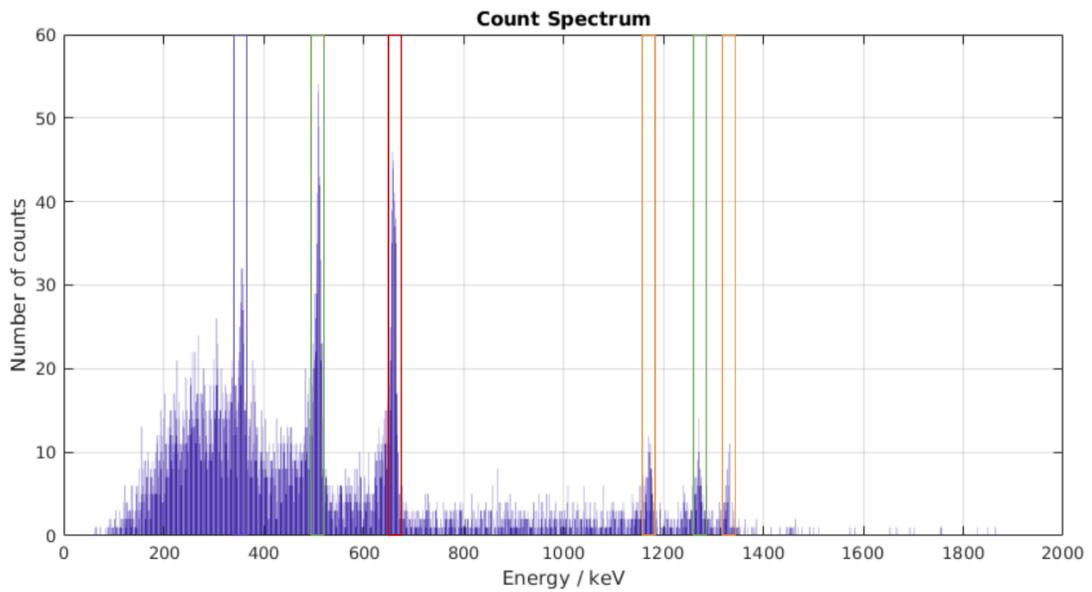

Figure 10 - The 20 keV energy windows used to localize sources in test 9 are shown as boxes with the following corresponding colors Ba-133 (purple), Cs-137 (red), Na-22 (green), and Co-60 (orange)

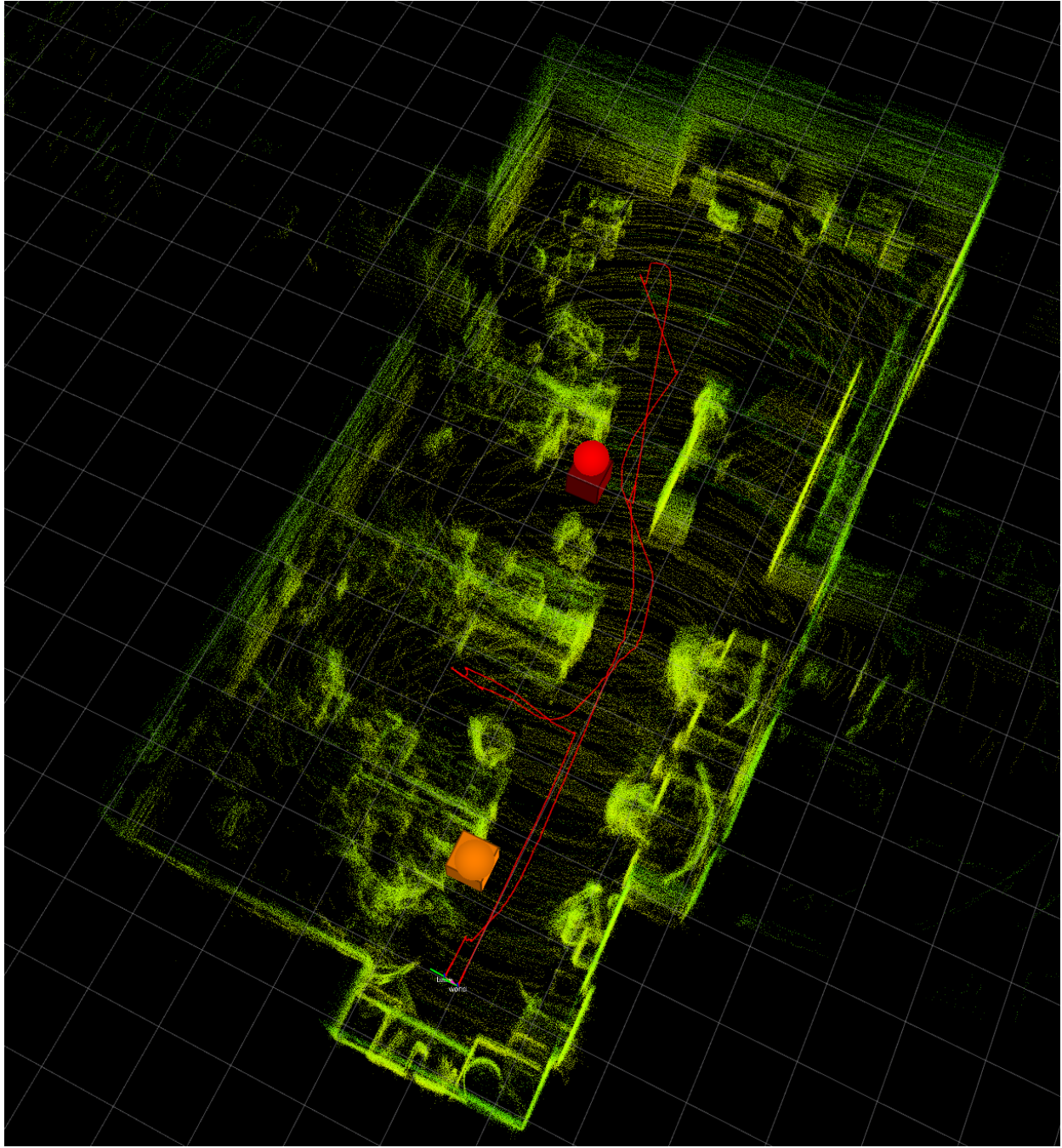

Figure 11 - Estimated locations of Co-60 (orange) and Na-22 (red) as cubes, with ground truth source location as spheres, trajectory of robot in red, and contextual spatial map in green (test 5).

**DISCUSSION**

The quantitative results in which the estimated source locations are directly measured against ground truth and the results in which multiple sources are clearly distinguished in two environments suggest this framework to be an accurate, efficient method of localizing multiple radiation sources. All sources were localized to within half a meter when compared to the ground truth across all tests, which sufficiently highlights key regions for possible further inspection. The high variance in the source localization accuracy can be attributed to the high sensitivity of radiation detection to many factors, including duration of dwell, proximity to source, and strength of source. Future experiments could isolate each factor individually and study its effect on the accuracy and the time required for localization.

A weakness in the presented approach was that the radiation data that was gathered in-situ but processed offline. Lack real-time feedback on the radiation image quality led to certain sources being insufficiently imaged, as noted previously. Thus another direction for future work is online radiation imaging. Various techniques have been employed to speed up the construction of the radiation map, such as the exclusion of voxels corresponding to free space from the radiation map as sources are found on surfaces or in containers [14], and the use of multiresolution maps that selectively focuses resources on regions of high source probability [17].

**CONCLUSIONS**

This work presents the integration of a self-localizing mobile robot with a Polaris-H Compton gamma camera for the detection and localization of radiation sources, supported by results from multiple laboratory experiments with varying source strengths, energies, and quantities. Tests based on discrete dwells and continuous motion were performed and the accuracies of the individual source locations estimated by the MLEM radiation imaging algorithm were compared against ground truth measurements. Mobile robot odometry was sufficiently accurate to allow for localization of sources to an average error of 0.2 m.

In the comings decades, robotics will begin to play an increasing role in the decommissioning industry. In the US, the Department of Energy estimates that the decommissioning of legacy nuclear facilities left behind from the Cold War alone will cost more than $240 billion through 2075 [18]. The preclusion of human entry in many of these facilities due to high levels of radiation and the general ALARA principle presents a unique opportunity for robots in the coming decades to assist in clean up and in gathering the intelligence needed to plan the decommissioning process. However, state-of-the-art radiation maps lack the ability to model the activity, uncertainty, and regional nature of real sources and planning algorithms lack the ability to leverage the map to follow a path that optimally and safely trades off the expected gains and cost in a nuclear setting. Future work will explore an efficient, risk-aware active perception framework for autonomous radiation mapping.